# Human-annotated label noise and their impact on ConvNets for remote sensing image scene classification

Longkang Peng, Tao Wei, Xuehong Chen, Xiaobei Chen, Rui Sun, Luoma Wan, Jin Chen, Xiaolin Zhu, *Senior Member, IEEE*

*Abstract*—Convolutional neural networks (ConvNets) have been successfully applied to satellite image scene classification. Human-labeled training datasets are essential for ConvNets to perform accurate classification. Errors in human-annotated training datasets are unavoidable due to the complexity of satellite images. However, the distribution of real-world human-annotated label noises on remote sensing images and their impact on ConvNets have not been investigated. To fill this research gap, this study, for the first time, collected real-world labels from 32 participants and explored how their annotated label noise affect three representative ConvNets (VGG16, GoogleNet, and ResNet-50) for remote sensing image scene classification. We found that: (1) human-annotated label noise exhibits significant class and instance dependence; (2) an additional 1% of human-annotated label noise in training data leads to 0.5% reduction in the overall accuracy of ConvNets classification; (3) the error pattern of ConvNet predictions was strongly correlated with that of participant's labels. To uncover the mechanism underlying the impact of human labeling errors on ConvNets, we further compared it with three types of simulated label noise: uniform noise, class-dependent noise and instance-dependent noise. Our results show that the impact of human-annotated label noise on ConvNets significantly differs from all three types of simulated label noise, while both class dependence and instance dependence contribute to the impact of human-annotated label noise on ConvNets. These observations necessitate a reevaluation of the handling of noisy labels, and we anticipate that our real-world label noise dataset would facilitate the future development and assessment of label-noise learning algorithms.

## I. INTRODUCTION

THE scene-level classification in remote sensing aims to assign a semantic category to a clipping patch of remote sensing image based on its content, which is a fundamental task in interpreting remote sensing images. This task significantly contributes to numerous real-world applications such as urban planning [1], land-use and land-cover mapping [2], [3], and climate monitoring [4]. Current deep learning algorithms, especially Convolutional Neural Networks (ConvNets), have achieved remarkable performance in this task [5], [6]. These methods generally require annotated data, consisting of input samples (i.e., images) along with their corresponding output labels (i.e., semantic categories), for adjusting network internal parameters (weights and biases), minimizing the discrepancy between network predictions and ground truth labels and then effectively improving network performance. The accuracy of training data thus significantly impacts the performance of ConvNets [7], [8]. Unfortunately, labels in training data are not always guaranteed to be accurate due to human annotator mistakes or ambiguities in the data [9], [10], [11], [12]. However, most literature in the remote sensing community assumes the human labelled data as the ground truth with accurate representation of reality, neglecting the inevitable noise in them [13]. Consequently, understanding the impact of unavoidable label noise on ConvNets is vital to improve their reliability in classifying remote sensing image scenes.

The label noise issue has been extensively studied in natural image recognition, which could provide insights into the impact of label noise in remote sensing image scene classification since both tasks aim to accurately classify the objects or scenes depicted in the images. To investigate the influence of label noise on ConvNets, studies in natural image recognition primarily vary noise rates of controllable simulated label noise. Results showed a relatively lower rate of growth in prediction errors caused by increased simulated label noise compared to the rate of growth in simulated label noise itself, suggesting a considerable resilience/robustness of ConvNets to such simulated label noise [14], [15], [16], [17], [18], [19], [20], [21], [22] but see [23]. However, the distribution of the simulated

This study was supported by Shenzhen Higher Institution Stability Support Plan (20200812154629001), Shenzhen Peacock Plan under Project (000517), the Otto Poon Charitable Foundation Smart Cities Research Institute at The Hong Kong Polytechnic University (Q-CDBP) and the Hong Kong Public Policy Research Funding Scheme (2022.A6.191.22A). *(Corresponding author: Tao Wei.)*

Longkang Peng and Xiaobei Chen are with the School of Psychology, Shenzhen University, Shenzhen 518060, China, and also with the Department of Land Surveying and Geo-Informatics, The Hong Kong Polytechnic University, Hong Kong, China (e-mail: longkang.peng@connect.polyu.hk; xiaobei.chen@connect.polyu.hk).

Tao Wei and Rui Sun are with the School of Psychology, Shenzhen University, Shenzhen 518060, China (e-mail: tao.wei@szu.edu.cn; 1586372849@qq.com).

Xuehong Chen and Jin Chen are with the State Key Laboratory of Remote Sensing Science, Faculty of Geographical Science, Beijing Normal University, Beijing 100875, China (e-mail: chenxuehong@bnu.edu.cn; chenjin@bnu.edu.cn).

Luoma Wan and Xiaolin Zhu are with the Department of Land Surveying and Geo-Informatics and the Otto Poon Charitable Foundation Smart Cities Research Institute, The Hong Kong Polytechnic University, Hong Kong, China (e-mail: luoma.wan@polyu.edu.hk; xiaolin.zhu@polyu.edu.hk).

noise is not generated based on real-world label noise, making it difficult to generalize the results to real-world label noise.

To address this issue, researchers attempted to explore the impact of real-world label noise in natural image recognition using the following two approaches. Jiang and his colleagues [24] simulated real-world label noise by replacing target images with incorrectly labeled web images while keeping labels unchanged. The results showed similar robustness of ConvNets with this type of label noise as with simulated label noise. Noteworthy, the way used to simulate real-world label noise involved human interventions in determining the target images for replacement, potentially introducing arbitrariness that could distort the distribution of noise. An alternative approach to investigate the impact of real-world label noise is to take advantage of human annotation errors in existing datasets. Van Horn et al. [25] measured the label accuracy of datasets CUB-200-2011 [26] and ImageNet [27], and found a reduction in accuracy due to noisy labels. However, these datasets have fixed noise rates and are limited in analyzing ConvNets against label noise that varies in real-world applications. Addressing this limitation, Wei et al. [28] constructed two benchmark datasets (CIFAR-10N and CIFAR-100N) with human-annotated real-world noisy labels at five noise rates. In particular, training images from datasets were divided into numbers of groups and then each group were annotated separately by 3 independent workers recruited from Amazon Mechanical Turk (MTurk). In this way, Wei et al. [28] showed quantitatively and qualitatively that human-annotated label noise was instance-dependent and found that human-annotated label noise had a more significant negative impact on ConvNets than simulated label noise at various noise rates.

Compared to natural images, remote sensing image scenes presented more orientation and scale variations [29], [30]. The abovementioned findings on natural image recognition datasets thus need to be further verified on remote sensing image scene datasets. To understand the influence of label noise on ConvNets, recent studies examined the performance of ConvNets on remote sensing image scene datasets with simulated label noise under different noise rates [7], [8]. In these experiments, a series of simulated label noises are injected into a well-labeled dataset, thus allowing for the controllable variation of noise rates in the dataset to reflect different magnitudes of label corruption encountered in real-world applications. Results revealed that the performance of ConvNets decreased as the noise rate increased and the magnitudes of this performance decrease were smaller compared to the increase in the noise rate, revealing a robustness of ConvNets to these simulated label noise [7], [8]. However, due to the lack of a remote sensing image scene dataset with real-world label noise, no study has thoroughly analyzed the property and impact of real-world label noise in the remote sensing image scene classification.

To fill this gap, we designed a series of human behavioral and computational experiments to quantitatively characterize the property of real-world label noise and systematically evaluate their influence on ConvNets in remote sensing image scene classification. In particular, we first collected real-world labels for the UCMerced dataset [31], a widely used remote sensing image scene dataset, from 32 participants, and quantitatively evaluated the pattern of human-annotated label noise. Then, these real-world labels with human-annotated label noise were employed to train three typical ConvNets: VGG16 [32], GoogLeNet [33], and ResNet-50 [34]. The generated ConvNets were subsequently utilized to analyze the impact of real-world label noise in the following three aspects: 1) assessing whether real-world label noise in training data significantly deteriorates the performance of ConvNets via overall accuracy analysis; 2) investigating whether the error patterns in the trained ConvNets resemble those of real-world label noise through error-pattern similarity analysis; and 3) determining whether ConvNets either amplify or tolerate real-world label noise in robustness analysis. Last, to uncover the mechanism of the impact of real-world label noise, we compared the impact of human-annotated label noise with three popular types of simulated label noise: uniform noise, class-dependent noise, and instance-dependent noise.

## II. MATERIALS AND EXPERIMENTAL DESIGN

### A. Dataset

UCMerced land use dataset [31] one of the most widely used datasets in remote sensing image scene classification [6], [35], [36], [37], [38], [39], [40], was used in this study. It consists of 21 land-use classes over different US regions and 100 aerial RGB images for each category. Each satellite image scene has 256 by 256 pixels with a pixel resolution of one foot. Among the 2100 aerial images, we selected 600 images from 3 artificial categories (airplane, freeway, and runway) and 3 nature categories (beach, forest, and river) for the experimental study (Fig. 1).

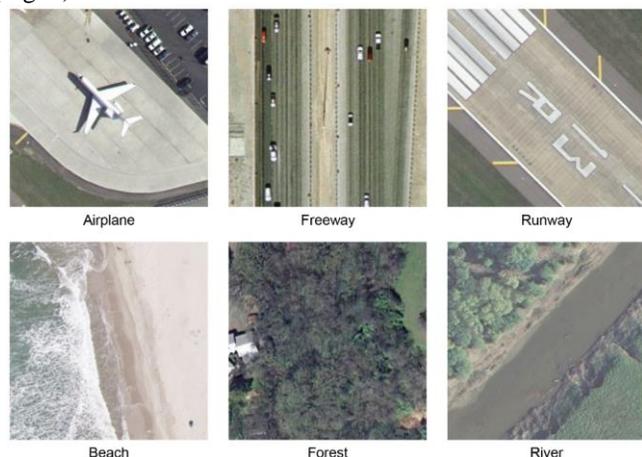

**Fig. 1.** Example images of six categories.

### B. Experimental Design

Our study includes three serials of experiments as shown in Fig. 2. First, in the behavioral experiment, real-world labels from participants were collected and analyzed to uncover the underlying pattern of human-annotated label noise. Then, in the assessment experiment, human labels were used to train ConvNet models, and the impact of human-annotated label noise on ConvNets was assessed in three aspects: 1) whether overall accuracy of ConvNets is affected by human-annotated label noise in training data (overall accuracy analysis); 2) whether the error pattern of ConvNets mirrors that of human-annotated label noise (error-pattern similarity analysis); and 3) whether ConvNets are resilient to human-annotated label noise (robustness analysis). Finally, in the comparative experiment, different ConvNet models were trained based on the simulated training data with the same mislabeling rate as human-annotated training data but with different class dependence or instance dependence of real-world noisy labels, and were then compared to reveal how different types of simulated label noise affect ConvNet classification performance.

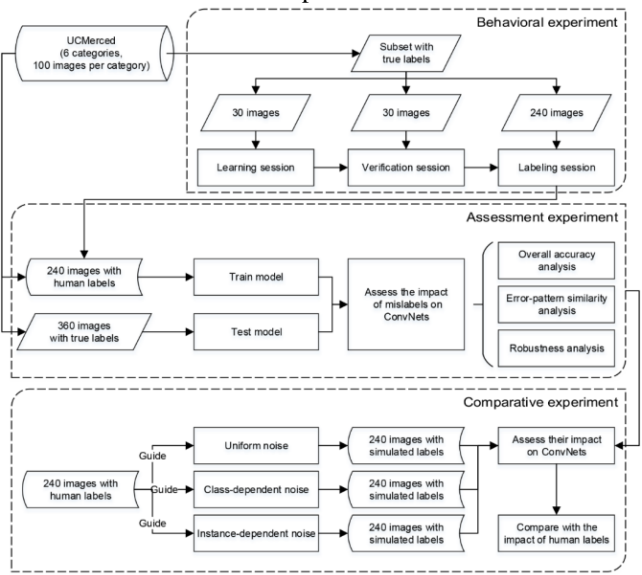

**Fig. 2.** Flowchart for the experiments.

1) *Behavioral Experiment:* In order to investigate the pattern and distribution of human-annotated label noise in remote sensing image scenes, we first conducted a behavioral experiment to obtain real-world labels on the UCMerced dataset from trained college students. The experimental protocol was approved by the Shenzhen University Institutional Review Board.

1a) *Participants:* 32 college students participated in the behavioral experiments (18 females; mean age 20). All participants were recruited from Shenzhen University and had normal or corrected-to-normal vision. They signed an informed consent form prior to the experiment and were paid for their participation afterward.

1b) *Stimulus:* From UCMerced, 300 remote sensing images were selected from six categories with 50 images per category. All images were randomly assigned to the learning, verification, and labeling sessions of 30 (5 images per category), 30 (5 images per category), and 240 (40 images per category) respectively.

1c) *Design and procedure:* Each experiment has three sessions: learning, verification, and labeling sessions. The first two sessions were completed on an online questionnaire platform (www.wjx.cn) and the labeling session was tested off-line on E-prime 3.0.

In the learning session, participants' task was to learn the relationship between remote sensing images and their category labels. On each trial, they were presented with an image with the category label. When they thought they had learned the relation, they pressed a button to proceed to the next trial. There was no time limit for learning. In this session, participants learned a total of 30 remote sensing images, five from one of six categories. The order of images was all randomized across participants.

In the verification session, participants were required to classify the remote sensing images using the category labels that they acquired in the learning session. On each trial, participants were presented with a remote sensing image with six category labels. If they chose the correct category label, they would move on to the next trial. Otherwise, the correct label would appear. Each trial had no time limit. There were 30 trials in total with five trials in each category. The purpose of this session was to give feedback to participants. If they were not satisfied with their performance, they could redo the learning session.

In the labeling session, participants were instructed to respond by pressing the corresponding key on the keyboard to classify the remote sensing images. A sheet of paper with six category labels and corresponding keys was placed on the desk to remind participants. As shown in Fig. 3, each trial began with a fixation point "+" in the center of the screen for 500ms. A remote sensing image followed immediately after the fixation point disappeared. Different from the other sessions, the image disappeared once participants pressed the key or it remained for 2000ms if participants did not press any keys. Then, a blank screen appeared for 500ms and the next trial appeared. There were 40 images from each of the six categories, resulting in a total of 240 trials. This session lasted approximately 20 minutes. Participants' key presses and response times were recorded by E-prime 3.0.

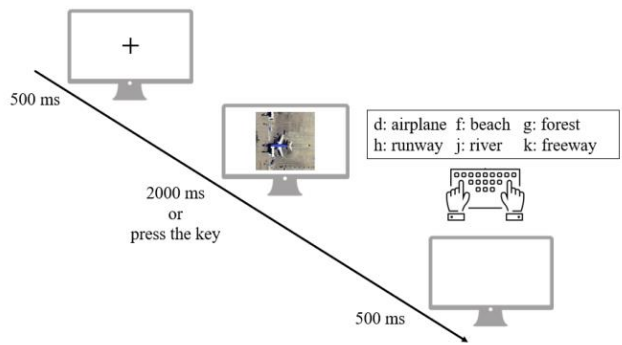

**Fig. 3.** The procedure of one trial in the labeling session.

1d) *Behavioral data analysis:* First, to evaluate whether human-annotated label noise were category-dependent (airplane, freeway, runway, beach, forest, river), the following generalized linear mixed effects analyses were conducted in the R (version 4.2.1) using lme4 (version 1.1-31; [41]) to explore the relationship between error rates and category. Because pictures were nested within categories, we included this nested structure as a random effect in addition to participants [41]. To test the main effect of category, we conducted likelihood ratio tests of the model with category as a fixed effect (full model) against the model without category (null model). To further test between which categories the error rates were significantly different, we conducted post-hoc contrast analyses with emmeans (version 1.8.5; [42]). Finally, to assess whether human-annotated label noise are instance-dependent, we examined whether participants were more prone to making errors when labeling a particular picture from a given category.

2) *Assessment Experiment:* The assessment experiment was designed to evaluate the impact of human-annotated label noise on CovnNet performance. Three typical ConvNets for scene classification, including VGG16 [32], GoogLeNet [33] and ResNet-50 [34], were selected for our experiment. These networks have been widely used as benchmarks for remote sensing image scene classification with good performance [35], [36], [37], [38], [39], [40] and represent the simple stacked structure, the Inception structure, and the residual learning structure, respectively. VGG16 has a simple architecture built by stacking multiple convolutional layers, max pooling layers and fully connected layers. GoogLeNet is built based on the Inception structure that adopts multiple different convolutional filter sizes in parallel to process visual information at various scales, resulting in a deep and wide architecture. ResNet-50 adopts residual learning structure performed by shortcut connections to construct significantly deeper networks compared to the VGG16 and GoogLeNet.

All the networks were implemented in PyTorch version 1.13.0 on an Ubuntu 9.4.0 system using 2 NVIDIA GeForce RTX 3090 GPUs. RAdam optimizer [43] with a learning rate of 10-4 was used as the optimization algorithm for training. The batch size was fixed at 128, and the standard cross-entropy was used as the loss function. The pretrained weights in ImageNet were used to initialize the network, which was then trained with the learning scheduler ReduceLROnPlateau. Human participants' responses to 240 images in the labeling session constituted the training set of ConvNets, and other 360 images from UCMerced (60 images from each of the six selected categories) formed the testing set to evaluate the performance of ConvNets with human label training. Specifically, every participant's responses in the labeling session were used to train three ConvNets, resulting in a total of 32 copies for each ConvNet. The trained copies of ConvNets were then evaluated with the testing set.

The impact of human-annotated label noise on ConvNets was assessed in three aspects. First, the overall accuracy (OA) (Stehman, 1997) was used to quantitatively measure the performance of ConvNets, which refers to the proportion of correctly classified instances to the total instances:

$$OA = \frac{\sum_{i=1}^{6} N_{ii}}{\sum_{i=1}^{6} N_i} \quad (1)$$

where $N_{ii}$ denotes the number of instances belonging to category $i$ and labeled as category $i$, and $N_i$ denotes the number of instances belonging to category $i$. The OA of ConvNets trained with human labels (Fig. 4 (a)) was compared with that of ConvNets trained with error-free labels (i.e., the original labels contained in the UCMerced dataset) to test whether the performance of ConvNets was affected by human-annotated label noise (Fig. 4 (b)).

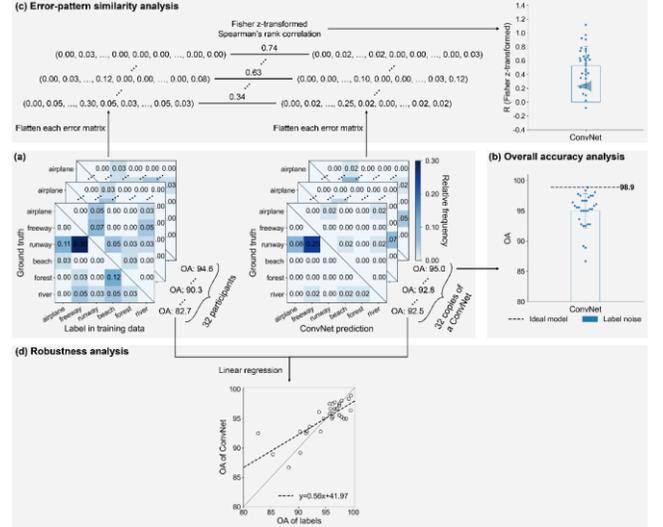

**Fig. 4.** Schematic diagram for assessing the impact of label noise on ConvNets. (a) The error patterns of 32 participants in the labeling session (left) and the error patterns of corresponding 32 copies of a ConvNet trained with human-annotated label noise (right). The number in each entry represents the relative frequency of incorrectly classified instances. (b) Overall accuracy analysis. One sample t-test was carried out for testing the null hypothesis that the mean of OA of ConvNets trained with human-annotated label noise is equal to the OA of ideal model trained with error-free training dataset. The dash line denotes OA of ideal model. The bar represents the mean and error bar represents the standard deviation of OA of ConvNets trained with label noise. Dots represent the OA of ConvNets trained with the dataset corresponding to each participant's annotated label noise. (c) Error-pattern similarity analysis. The error matrix of one participant and its corresponding ConvNet were flattened and then the Spearman's rank correlation between these two vectors was calculated. This procedure was repeated for every participant, resulting in a total of 32 Spearman's rank correlation coefficients (ρs). These ρs were Fisher's z-transformed and shown as dots. The bar represents the mean and error bar represents the standard deviation of Fisher's z-transformed ρs. The distribution represents the probability density function of Fisher's z-transformed ρs in permuation tests by randomly pairing human error matrix and ConvNet error matrix 1000 times. (d) Robustness analysis. A linear regression model, indicated by the dash line, was built between the OA of 32 participants and the corresponding ConvNets.

Then, representational similarity analyses (RSA, [44]) were conducted to assess whether the error pattern exhibited by a ConvNet reflects the error pattern present in its training data derived from a participant (Fig. 4 (c)). First, the unique pattern of labeling errors made by each participant in the labeling session can be represented by an error matrix as shown in Fig. 4 (a). Each non-diagonal element in the matrix is a specific type of error made by the participant, which is relative error frequency calculated as:

$$F_{ij} = \frac{N_{ij}}{N_i}, i \neq j \quad (2)$$

where $F_{ij}$ and $N_{ij}$ are the relative frequency and the number of instances belonging to category $i$ but mislabeled as category $j$, respectively, and $N_i$ is the number of instances belonging to category $i$. The labels created by each participant were used to train ConvNet model and that trained model obtained its error matrix from the test dataset as shown in Fig. 4 (a). To assess the relation between human labeling errors and corresponding ConvNet errors, the error matrix of one participant and its corresponding ConvNet matrix were flattened and then a Spearman's rank correlation coefficient between these two vectors was calculated and Fisher's-z transformed, given that the flatted error matrices did not fulfill the assumptions required by Pearson's correlation coefficient. The significance was tested by permutation tests by randomly pairing human error matrix and ConvNet error matrix 1000 times.

To answer how well a ConvNet performed under different noise levels, robustness analysis was conducted. As shown in Fig. 4 (d), the OA of 32 participants and the corresponding ConvNets was generated from their error matrices (Fig. 4 (d) and (e)) and then a linear regression model was built between these two groups of OA. A smooth linear model with a slope less than 1 indicates that the ConvNet is relatively robust to the label noise, and conversely that the ConvNet amplifies the label noise.

3) *Comparative Experiment:* To elucidate the mechanism underlying the impact of real-world label noise on ConvNets, we compared the effect of human-annotated label noise with three types of simulated label noise: uniform noise, class-dependent noise and instance-dependent noise. Compared to real-world label noise, uniform noise neglects the class and instance dependence, class-dependent noise neglects the instance dependence, and instance-dependent noise neglects the class dependence. Thus, the comparison between the impact of real-world label noise and these simulated label noise provided an insight into the contribution of class and instance dependence in the impact of real-world label noise.

These simulated labels with label noise were generated for the 240 images in the labeling session of the behavioral experiment (section 2.2.1). Based on every participant's labeling error (Fig. 5 (a)), the uniform noise is generated by randomly selecting the same number of instances as participant' mislabels and replacing correct labels with any other labels with a uniform probability. As a result, the uniform noise is both instance-independent and class-independent (Fig. 5 (c) and (d)). Class-dependent noise is generated by randomly relocating mislabels of a particular type of error from participants to any instances within the same class (Fig. 5 (f)). Thus, the class-dependent noise is instance-independent but maintains the same pattern with participants' labeling errors (Fig. 5 (e)). Instance-dependent noise is generated by randomly replacing mislabels of instances from participants (Fig. 5 (b)) with any mislabels with a uniform probability (Fig. 5 (h)). Therefore, the instance-dependent noise is class-independent but holds that those instances mislabeled by participants are still incorrectly labeled (Fig. 5 (g)). These three types of simulated label noise were repeated 30 times for each participant to consider the random effect of sample selection and label flipping. We trained the three ConvNets with these simulated labels and assessed their performance in the testing set as described in Section 2.2.2. Similar to the assessment experiment, we conducted overall accuracy, error-pattern similarity and robustness analyses for all simulations. Finally, the outputs from 30 repetitions associated with each participant's error distribution were averaged and then compared between three simulated label noise and human-annotated label noise to inform the difference in the impact of simulated and human noisy labels on ConvNets.

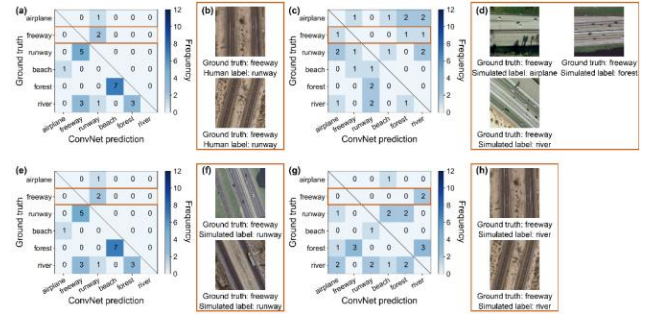

**Fig. 5.** Illustration of three types of simulated label noise. (a) The error pattern of one participant. The noise patterns of (c) uniform noise, (e) class-dependent noise and (g) instance-dependent noise, all of which are modeled based on (a). The number in each entry represents the frequency of incorrectly classified instances. Freeway images with mislabels (orange boxes in (a), (c), (e) and (g)) in (b) human labels, (d) uniform noise, (f) class-dependent noise and (h) instance-dependent noise.

## III. RESULTS

### A. Behavior Experiment: Noise Pattern of Human-annotated Label Noise

Fig. 6 shows the error rates in each category and Table I shows the results of generalized linear mixed effects analyses with category as a fixed effect. Adding Category to the model significantly improved Model fitting ($p < .001$), suggesting that participants made more errors in some categories than others. Post-hoc contrast analyses further showed that the error rate of airplane is lower than all categories except beach, and the error rate of beach is lower than freeway and river (Table II). There are no significant differences between other categories.

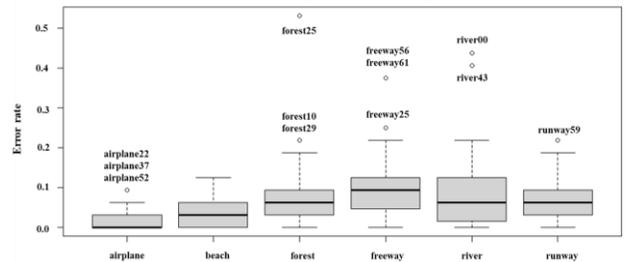

**Fig. 6.** The boxplots of human labeling error rates in each category.

To investigate whether participants' performance was affected by specific pictures, we tried to identify outliers. As shown in Fig. 6, outliers were identified for all categories except for beach, suggesting that participants were more likely to make error on some pictures than others, i.e., human mislabels are also instance-dependent. In sum, participants' errors were both category- and instance-dependent, demonstrating the errors made by them were not random.

TABLE I
THE RESULT OF MODEL COMPARISON

| Model | Random effect | Fixed effect | AIC | BIC | LogLik | Dev | X² | df | p |
|---|---|---|---|---|---|---|---|---|---|
| Null model | (1|Subject) | - | 3561.7 | 3582.6 | -1777.9 | 3555.7 | | | |
| Full model | +(1|Category/Picture) | Category | 3416.9 | 3479.4 | -1699.5 | 3398.9 | 156.85 | 6 | < 2.2e-16*** |

* $p < .05$, ** $p < .01$, *** $p < .001$

TABLE II
THE RESULT OF POST HOC MULTIPLE COMPARISONS

| Contrast | b | SE | z | p |
|---|---|---|---|---|
| airplane-beach | -0.58 | 0.29 | -1.99 | .35 |
| airplane-forest | -1.29 | 0.28 | -4.61 | .0001*** |
| airplane-freeway | -1.63 | 0.27 | -5.94 | <.0001*** |
| airplane-river | -1.36 | 0.28 | -4.90 | <.0001*** |
| airplane-runway | -1.21 | 0.28 | -4.33 | .0002*** |
| beach-forest | -0.70 | 0.26 | -2.76 | .06 |
| beach-freeway | -1.05 | 0.25 | -4.19 | .0004*** |
| beach-river | -0.78 | 0.25 | -3.07 | .03* |
| beach-runway | -0.63 | 0.26 | -2.46 | .14 |
| forest-freeway | -0.34 | 0.23 | -1.48 | .68 |
| forest-river | -0.08 | 0.24 | -0.32 | .99 |
| forest-runway | 0.07 | 0.24 | 0.309 | .99 |
| freeway-river | 0.27 | 0.23 | 1.156 | .86 |
| freeway-runway | 0.42 | 0.23 | 1.787 | .47 |
| river-runway | 0.15 | 0.24 | 0.629 | .99 |

* $p < .05$, ** $p < .01$, *** $p < .001$ (the p-value of post hoc multiple comparisons was adjusted by Tukey test)

*B. Assessment Experiment: Impact of Human-annotated Label Noise on ConvNets*

1) *Overall Accuracy Analysis:* In comparison with the ideal model trained with an error-free training dataset (dash lines in Fig. 7), ConvNets trained with human-annotated label noise (red bars in Fig. 7) show significantly lower overall accuracy (one-sample t-test: VGG16, $t_{(31)} = -7.719$, $p < .0001$; GoogLeNet, $t_{(31)} = -5.604$, $p < .0001$; ResNet-50, $t_{(31)} = -4.607$, $p < .0001$), suggesting that mislabels of human annotation have a significant negative impact on the performance of ConvNets. In addition, the impacts are different across ConvNets (one-way ANOVA: $F_{(2,93)} = 12.446$, $p < 0.0001$): greater impact on degrading performance of VGG16 than GoogLeNet and ResNet-50 (red bars in Fig. 7; LSD multiple comparison test: VGG16 vs. GoogLeNet, $p < .0001$; VGG16 vs. ResNet-50, $p < .0001$; GoogLeNet vs. ResNet-50, $p = .808$).

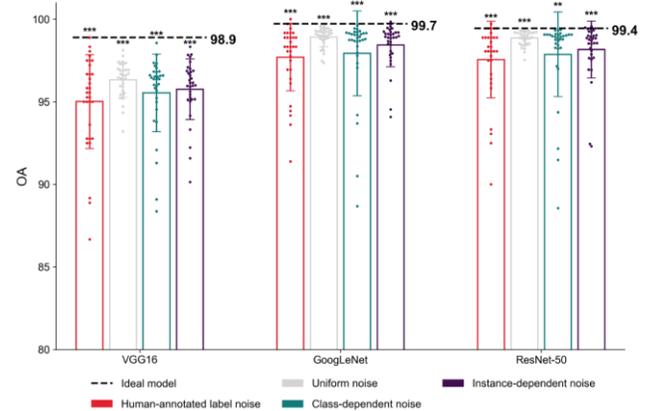

**Fig. 7.** Overall accuracy (OA) of VGG16, GoogLeNet and ResNet-50 trained with human-annotated and simulated label noise. The dash lines denote OAs of ideal models trained with error-free training dataset. Bars represent the mean OA of ConvNets trained with label noise. Dots represent the OA of ConvNets trained with the dataset corresponding to each participant's annotated and simulated label noise (one dot from ConvNets trained with simulated label noise represents the mean OA of 30 simulations associated with one participant's error distribution). Error bars represent the standard deviation of OA of ConvNets trained with label noise. Asterisks denote significant difference between the OA of ConvNets trained with label noise and those trained with an error-free dataset, *$p < .05$, **$p < .01$, ***$p < .001$.

2) *Error-pattern Similarity Analysis:* Significant error-pattern similarity was observed between human and ConvNets error matrices (red bars in Fig. 8). Permutation tests were conducted 1000 times to generate the distribution of Spearman's rank correlation coefficients between randomly paired human error matrix and ConvNet error matrix (distributions in Fig. 8). Fisher's z-transformed spearman's rank correlation coefficients between human error matrix and ConvNet error matrix were significantly higher than randomized pairs (VGG16, $p < .0001$; GoogLeNet, $p < .0001$; ResNet-50, $p < .0001$), demonstrating that the errors made by the trained ConvNets stem from specific individual training dataset.

Moreover, a one-way repeated measure ANOVA on Fisher's z-tranformed correlation coefficients of these models showed that error-pattern similraity differed across ConvNets (Fig. 8, $F_{(1.687, 52.300)} = 3.845$, $p < .05$, Greenhouse-Geisser-corrected). Post-hoc analysis with a LSD adjustment revealed that VGG16 exhibited significantly stronger similarity than GoogLeNet ($p < .01$), and there was no significant difference in similarities between VGG16 and ResNet-50 ($p = .120$) and in similarities between GoogLeNet and ResNet-50 ($p = .391$).

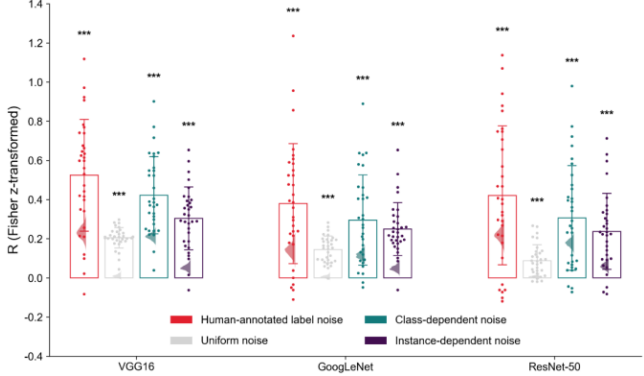

**Fig. 8.** Fisher's z-transformed Spearman's rank correlation coefficients (ρs) between the error matrix of a training dataset and error matrix of its corresponding trained model output across three CNN models under different noise conditions. Bars represent the mean and error bars represents the standard devidation of Fisher's z-transformed ρs of ConvNets trained with label noise. Dots represent the Fisher's z-transformed ρs of ConvNets trained with the dataset corresponding to each participant's annotated and simulated label noise (one dot from ConvNets trained with simulated label noise represents the mean Fisher's z-transformed ρ of 30 simulations associated with one participant's error distribution). The distributions represent the probability density function of Fisher's z transformed ρs in permuation tests by randomly pairing human error matrix and ConvNet error matrix 1000 times. Asterisks denote the significance of permutation tests, * $p < .05$, ** $p < .01$, *** $p < .001$.

3) *Robustness Analysis:* Human-annotated label noise was not amplified by ConvNets, although it significantly affects ConvNets' overall accuracy and error pattern. As shown in Fig. 9(a), when the overall accuracy of human labels decreased by 1%, the overall accuracy of ConvNets decreased by around 0.5% (0.56% for VGG16, 0.42% for GoogLeNet, and 0.46% for ResNet-50), demonstrating ConvNets' robustness to human-annotated label noise.

### C. Comparative Experiment: Comparison of Human-annotated Label Noise and Simulated Label Noise

To uncover the mechanism of the impact of human-annotated label noise, we added simulated uniform noise, class-dependent noise, and instance-dependent noise to the training dataset and investigated how these types of noise affect ConvNets in comparison with human-annotated label noise.

1) *Overall Accuracy Analysis:* The results of overall accuary analysis are shown in Fig. 7. All ConvNets trained with

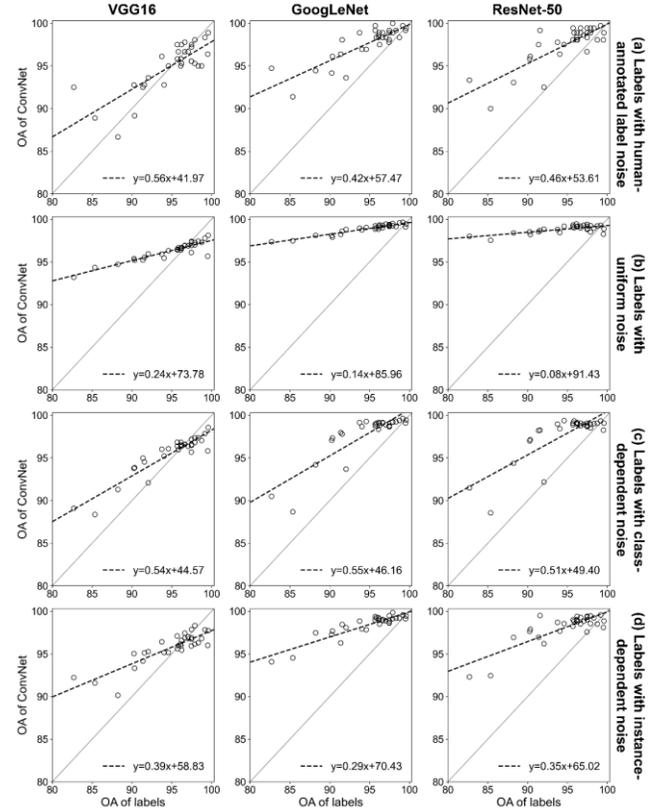

**Fig. 9.** Associations between the OA of training datasets with label noise, i.e., (a) human annotated label noise, (b) uniform noise, (c) class-dependent noise, and (d) instance-dependent noise, and the OA of VGG16, GoogLeNet and ResNets trained using these datasets. Note that one dot from ConvNets trained with simulated label noise represents the mean OA of 30 simulations associated with one participant's error distribution.

simulated label noise significantly performed worse than the ideal models trained with error-free training dataset (Fig. 7, one-sample t-tests: $t$s < -3.482, $p$s < .01). To compare the impacts of different label noise across different ConvNets, a 4 (noise types: human-annotated, uniform, class-dependent, instance-dependent) by 3 (ConvNet types: VGG16, GoogLeNet, ResNet-50) two-way repeated measures ANOVA was conducted. The results revealed a significant main effect of noise type ($F_{(1.654, 51.276)} = 9.858$, $p < .001$, Greenhouse-Geisser-corrected). Planned contrasts showed that compared with human-annotated label noise, all simulated noises have a weaker impact on overall accuracy (class-dependent noise, $p = .084$; instance-dependent and uniform noise, $p$s < .001). Among all noises, uniform noise has the weakest impact on ConvNets' performance (uniform vs. human-annotated, $p < .001$; uniform vs. class-dependent, $p < .05$; uniform vs. instance-dependent, $p < .01$). The main effects of ConvNet types was also significant ($F_{(1.302, 40.363)} = 225.798$, $p < .0001$, Greenhouse-Geisser-corrected), with VGG16 reporting significantly lower overall accuracy compared to GoogLeNet and ResNet-50 (VGG16 vs. GoogLeNet, $p < .0001$; VGG16 vs. ResNet-50, $p < .0001$; GoogLeNet vs. ResNet-50 $p = .098$). There was no significant interaction between noise types and ConvNet types on the overall accuracy of ConvNets ($F_{(2.517, 78.029)} = 0.673$, $p = .546$, Greenhouse-Geisser-corrected), suggesting that noise types affect different ConvNets in a similar way.

2) *Error-pattern Similarity Analysis:* As shown in Fig. 8, error patterns of ConvNets trained with both simulated and human-annotated label noise were similar to those of label noise itself and these similarities varied for different types of label noise and ConvNets. Across all ConvNets, significant similarity between error patterns of simulated noisy labels and ConvNets was observed by permutation tests ($p$s < .0001). To further compare the similarities of different label noise across ConvNets, a 4 (noise types: human-annotated, uniform, class-dependent, instance-dependent) by 3 (ConvNet types: VGG16, GoogLeNet, ResNet-50) two-way repeated measures ANOVA was conducted. The main effect of noise type was significant ($F_{(1.837, 56.948)} = 29.566$, $p < 0.0001$, Greenhouse-Geisser-corrected). Planned contrasts demonstrate that error-pattern similarity for human-annotated label noise was significantly higher than that for simulated label noise (human-annotated vs. uniform, $p < .0001$; human-annotated vs. class-dependent, $p < .001$; human-annotated vs. instance-dependent, $p < .001$). Among all simulated label noises, error-pattern similarity for uniform noise was significantly lower than that for instance-dependent noise (uniform vs. instance-dependent, $p < .0001$), and both were significantly lower than that for class-dependent noise (uniform vs. class-dependent, $p < .0001$; instance-dependent vs. class-dependent, $p < .05$). These results suggest that both class-dependence and instance-dependence of human-annotated label noise contribute to the similarity between error patterns of labels and ConvNets, making error patterns of ConvNets more similar to those of human-annotated label noise than simulated label noise.

The main effect of ConvNet type is also significant ($F_{(1.620, 50.217)} = 15.069$, $p < .0001$, Greenhouse-Geisser-corrected). Error-pattern similarity between noisy labels and VGG16 predictions was significantly stronger than that between noisy labels and the predictions of the other two CovnNets (VGG16 vs. GoogLeNet $p < .0001$, VGG16 vs. ResNet-50 $p < .001$, GoogLeNet vs. ResNet-50 $p = .828$). Noise type and ConvNet type affect error-pattern indepently, as shown by insignificant interaction between the noise types and ConvNet types on error-pattern similarity ($F_{(2.798, 86.746)} = 1.821$, $p = .153$, Greenhouse-Geisser-corrected).

3) *Robustness Analysis:* As demonstrated in Fig. 9 (b-d), 1% decrease in label accuracy due to simulated label noise reduced the average overall accuracy of ConvNets by around 0.5% or less, indicating robustness of ConvNets to simulated label noise. In comparison with ConvNets under human-annotated label noise (dash lines in Fig. 10), all ConvNets under simulated noises exhibited significantly different robustness: GoogLeNet and ResNet-50 under class-dependent noise exhibited significantly weaker robustness (GoogLeNet, $t_{(29)} = 7.548$, $p < .0001$; ResNet-50, $t_{(29)} = 4.336$, $p < .001$), and all other ConvNets with stronger robustness (one-sample t-tests: $t$s < -9.623, $p$s < .0001).

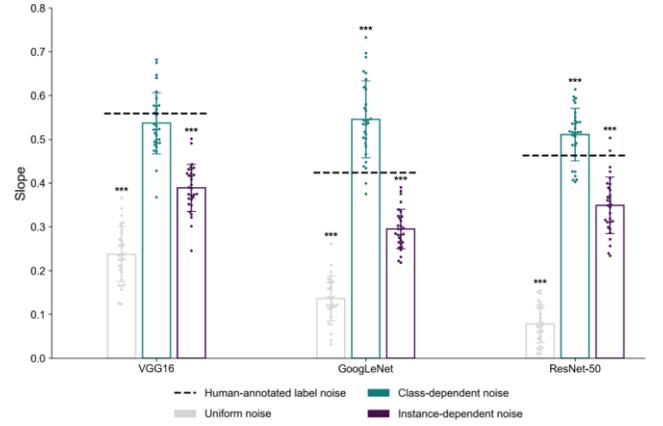

**Fig. 10.** Robustness of VGG16, GoogLeNet and ResNet-50 under the human-annotated label noise and the three types of simulated label noise. The dash lines denote the slopes of the linear regression between the OA of human labels against the OA of the corresponding ConvNets. Bars represent the means and errors bars represent the standard deviation of slopes of ConvNets trained with simulated label noise. Dots represent the slope of ConvNets trained with the dataset corresponding to each participant's simulated label noise (one dot represents the mean slope of 30 simulations associated with one participant's error distribution). Asterisks denote significant difference between the slopes of ConvNets trained with simulated label noise and those trained with human-annotated label noise, * $p < .05$, ** $p < .01$, *** $p < .001$.

To further compare the robustness of ConvNets under different simulated noise types, a 3 (noise types: uniform, class-dependent, instance-dependent) by 3 (ConvNet types: VGG16, GoogLeNet, ResNet-50) two-way repeated measures ANOVA was conducted. The main effect of simulated noise types is significant ($F_{(1.666, 48.319)} = 468.970$, $p < .0001$, Greenhouse-Geisser-corrected): all ConvNets were most robustness to uniform noise, followed by instance-dependent noise, and least robustness to class-dependent noise ($p$s < .0001, see Fig. 10). In addition, the interaction between simulated noise types and ConvNets is significant ($F_{(4, 116)} = 25.258$, $p < .0001$), suggesting that ConvNets' robustness to noises varies across different noise types. In the case of uniform noise, ResNet-50 exhibited the highest robustness, then GoogLeNet, while VGG16 demonstrated the least robustness ($p$s < .0001). For class-dependent noise, only ResNet-50 showed stronger robustness than VGG16 and GoogLeNet ($p$s < .05). Regrading instance-dependent noise, GoogLeNet demonstrated the most robustness, followed by ResNet-50, with VGG16 showing the least robustness ($p$s < .05).

### III. DISCUSSION

Our study quantitatively examined the property of real-world label noise, deepening our understanding of human labeling behaviors on remote sensing images. Previous works on label noise often have various assumptions on the simulated noise type ranging from class-dependent noise [45], [46], [47], [48] to instance-dependent noise [20], [49], [50]. In a recent investigation where each image was annotated by three MTurk workers, Wei and his colleagues [28] showed that real-world human annotation noisy labels follow an instance-dependent pattern rather than a class-dependent one. With images annotated by 32 participants, our findings provide statistical evidence that human-annotated label noise occurs more on certain categories and specific images compared to others, suggesting that real-world noisy labels exhibit both class- and instance-dependent patterns. Consequently, we propose to explore the issue of label-noise learning under the assumption that the label noise is dependent on both the class and the instance. By considering this dual pattern, we can gain further insights into effectively addressing label noise in real-world scenarios.

Compared to previous research on the issue of label noise, our study provides new insights into how ConvNets capture real-world noisy labels. Specifically, we found that human-annotated noisy labels significantly affect the error pattern of ConvNet predictions, extending beyond affecting the classification accuracy [28]. The similarity of the error pattern between human labels and the ConvNet predictions in our study suggests that the error pattern of ConvNet predictions could be used to represent the error pattern of noisy labels, which will aid the estimation of the noise transition matrix. The noise transition matrix plays a central role in the label-noise learning to improve the classification accuracy [51], [52], but it is generally unidentifiable in real-world applications due to the unknown error pattern of noisy labels. Additionally, the accumulation of real-world labels from a substantial number of participants provided our study a unique opportunity to analyze how robust the ConvNet is to real-world noisy labels. We observed that ConvNets exhibit some robustness to real-world label noise, beyond just simulated label noise [15]. However, we observed that a performance gap still exists between ConvNets trained with noisy labels and those trained with clean labels. It is worth further improving the robustness of ConvNets to bridge this gap by label-noise learning algorithms, such as SOP [53] and SN [54]. The availability of our dataset with real-world noisy labels (https://github.com/LK-Peng/Learning-with-Real-world-Label-Noise) could be adopted to assess existing label-noise learning algorithms and to foster the future development of these algorithms.

Our study presents a comprehensive comparison between human-annotated and simulated label noise and their impact on ConvNets in remote sensing image scene classification. The results showed that human-annotated label noise led to a significantly greater impact on both the overall accuracy and the error pattern of ConvNets than simulated label noise. However, the robustness of ConvNets to both types of label noise varies depending on the specific case. Although human-annotated label noise had a weaker impact on GoogLeNet and ResNet-50 compared to class-dependent noise (Fig. 10), this trend was reversed in certain classes, such as beach and river (Fig. A1). Future studies on label noise should focus on real-world human-annotated noisy labels to gain a comprehensive understanding of their outcome. The results further revealed that human-annotated label noise, as well as class- and instance-dependent noise, had a significantly greater impact on ConvNets than uniform noise. This suggests that both the class and instance dependence of human mislabels contribute to the impact of human-annotated label noise on ConvNets. Thus, we recommend building datasets that consider both the class and instance dependence of label noise when simulating noisy labels and developing practical label-noise learning algorithms capable of handling real-world label noise. However, obtaining real class and instance dependence poses challenges, particularly when the classification scheme varies across different tasks. While it is possible to build a large dataset with label noise from the web for natural scene images [24], obtaining web labels for remote sensing scene images is challenging due to the need for domain knowledge-based preprocessing. Conducting behavioral experiments to collect real-world labels is both costly and time-consuming, especially for large datasets. An alternative approach could be to collect human annotations for a small-scale dataset. This dataset would be used to fine-tune a teacher ConvNet. The outputs of this teacher ConvNet would have a similar error pattern to real-world human-annotated noisy labels due to the significant similarity between the error pattern of ConvNet predictions and human-annotated noisy labels. By feeding a substantial volume of data into the network and obtaining corresponding outputs, these input-output pairs can serve as a proxy for a large-scale dataset with real-world label noise, which can be used to explore label-noise issues on a large scale. Moreover, our study examined the impac of label noise on ConvNets training. However, it is also meaningful to evaluate the effect of labeling errors on ConvNets testing, as it is crucial for selecting a ConvNet in real-world deployment based on its test accuracy [55].

By comparing three representative ConvNets, our study provides insights into ConvNets structures. Among these three ConvNets, VGG16 had the lowest tolerance for human-annotated label noise as well as all three types of simulated label noise. Given VGG16's suboptimal performance in fine-tuning using ImageNet architectures compared to the other two ConvNets [34], we speculate that a ConvNet with a worse-pretrained architecture is likely to perform worse when confronted with noisy training labels. This aligns with findings on label noise from the web by [24]. In addition, considering VGG16's relatively shallow structure without branches, we hypothesize that the inclusion of a deeper structure, a branch structure, or a combination of both may enhance the network's ability to handle label noise. This suggests the need for future research on the effectiveness of various ConvNet structures in managing human-annotated label noise.

## IV. Conclusion

We obtained human-annotated noisy labels aimed at examining the property of real-world label noise and their impact on ConvNets for remote sensing scene classification. We qualitatively revealed that human-annotated label noise was dependent on both classes and instances. When training ConvNets with these human annotations, human mislabels significantly affected the classification accuracy and the error pattern of ConvNets while ConvNets demonstrated a certain degree of resilience towards these mislabels. We then statistically compare the impact of label noise on the classification accuracy, the error pattern, and the robustness of ConvNets when learning with human-annotated label noise and simulated label noise (i.e., uniform, class-dependent, and instance-dependent noises). Experiments demonstrated that the human-annotated label noise generally exhibited a larger impact on ConvNets' performance in terms of classification accuracy and error pattern compared to all simulated label noise while the degree of robustness of ConvNets to both types of label noise is case-by-case. Our study suggests the necessity of collecting real-world labels of remote sensing data and investigating their impact on various remote sensing classification tasks.

APPENDIX

Regarding images for each category (Fig. A1), the robustness of all ConvNets under human-annotated label noise was significantly weaker than class-dependent noise in images of the beach and the river (one-sample t-tests: $t$s $< -2.359$, $p$s $< .05$) and significantly stronger than class-dependent noise in images of runway (one-sample t-tests: $t$s $> 2.798$, $p$s $< .01$), demonstrating that the difference between the robustness of ConvNets to these two types of label noise was associated with the category of images.

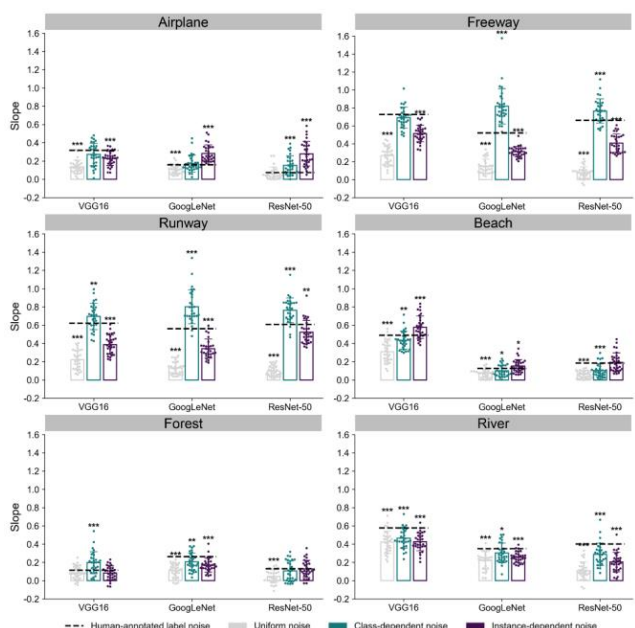

**Fig. A1.** Robustness of VGG16, GoogLeNet and ResNet-50 under the human-annotated label noise and the three types of simulated label noise for each category. The dash lines denote the slopes of the linear regression between the F1 score of human labels against the F1 score of the corresponding ConvNets. Bars represent the mean of slopes of ConvNets trained with simulated label noise. Dots represent the slope of ConvNets trained with the dataset corresponding to each participant's simulated label noise (one dot represents the mean slope of 30 simulations associated with one participant's error distribution). Asterisks denote significant difference between the slopes of ConvNets trained with simulated label noise and those trained with human-annotated label noise, $*p < .05$, $**p < .01$, $***p < .001$.